\documentclass[10pt,twocolumn,letterpaper]{article}

\usepackage{iccv}
\usepackage{times}
\usepackage{epsfig}
\usepackage{graphicx}
\usepackage{amsmath}
\usepackage{amssymb}

\usepackage{booktabs}

\usepackage{url}            
\usepackage{booktabs}       
\usepackage{amsfonts}       
\usepackage{nicefrac}       
\usepackage{microtype}      
\usepackage{xcolor}         
\usepackage{subcaption}
\makeatletter
\@namedef{ver@everyshi.sty}{}
\makeatother
\usepackage{tikz}
\usepackage{algorithm}
\usepackage{algorithmic}
\usepackage{amsthm}
\usepackage[shortlabels,inline]{enumitem}
\usepackage{wrapfig}
\newtheorem{definition}{Definition}
\newtheorem{theorem}{Theorem}[section]

\newtheorem{lemma}[theorem]{Lemma}
\definecolor{xablue}{HTML}{00ABF5}
\definecolor{xayellow}{HTML}{FACD00}
\definecolor{xapink}{HTML}{FF6D8A}
\definecolor{xagreen}{HTML}{00DDCB}
\definecolor{xagray}{HTML}{545454}

\usepackage{array}
\newcolumntype{H}{>{\setbox0=\hbox\bgroup}c<{\egroup}@{}}
\usepackage{multirow}

\usepackage{xr}
\makeatletter
\newcommand*{\addFileDependency}[1]{
  \typeout{(#1)}
  \@addtofilelist{#1}
  \IfFileExists{#1}{}{\typeout{No file #1.}}
}
\makeatother
\newcommand*{\myexternaldocument}[1]{%
    \externaldocument{#1}%
    \addFileDependency{#1.tex}%
    \addFileDependency{#1.aux}%
}
\myexternaldocument{mixpath_iccv23_supp}

\usepackage[pagebackref=true,breaklinks=true,letterpaper=true,colorlinks,bookmarks=false]{hyperref}

 \iccvfinalcopy 

\usepackage[capitalize]{cleveref}
\crefname{section}{Sec.}{Secs.}
\Crefname{section}{Section}{Sections}
\Crefname{table}{Table}{Tables}
\crefname{table}{Tab.}{Tabs.}

\def\iccvPaperID{6030} 

\ificcvfinal\fi

\begin{document}

\title{MixPath: A Unified Approach for One-shot Neural Architecture Search}

\author{
	Xiangxiang Chu \quad Shun Lu \quad Xudong Li \quad Bo Zhang \\
	{\small{cxxgtxy@gmail.com, lushun19s@ict.ac.cn, lixudong16@mails.ucas.edu.cn, zhangboyd@qq.com} 
	}
}

\maketitle
\ificcvfinal\thispagestyle{empty}\fi

\begin{abstract}
Blending multiple convolutional kernels is proved advantageous in neural architecture design. However, current two-stage neural architecture search methods are mainly limited to single-path search spaces. How to efficiently search models of multi-path structures remains a difficult problem. In this paper, we are motivated to train a one-shot multi-path supernet to accurately evaluate the candidate architectures. Specifically, we discover that in the studied search spaces, feature vectors summed from multiple paths are nearly multiples of those from a single path. Such disparity perturbs the supernet training and its ranking ability. Therefore, we propose a novel mechanism called \emph{Shadow Batch Normalization} (SBN) to regularize the disparate feature statistics. Extensive experiments prove that SBNs are capable of stabilizing the optimization and improving ranking performance. We call our unified multi-path one-shot approach as MixPath, which generates a series of models that achieve state-of-the-art results on ImageNet.
\end{abstract}

\section{Introduction}
Complete automation in neural network design is one of the most important research directions of automated machine learning \cite{tan2018mnasnet,liu2018darts}. Among various mainstream paradigms, one-shot approaches \cite{guo2019single,chu2019fairnas,li2019blockwisely} make use of a weight-sharing mechanism that reduces a large amount of computational cost, but these approaches mainly focus on searching for single-path networks. Observing that a multi-path structure is beneficial for model performance as in Inception \cite{szegedy2015going}  and ResNeXT \cite{xie2017aggregated}, it is necessary to incorporate multi-path structure into the search space. Noticeably, exploring multi-path search space is made possible in one-stage approaches like \cite{chu2019fair} and \cite{pham2018efficient}. However, it poses a challenge for two-stage methods  to train a one-shot supernet that can accurately predict the performance of its multi-path submodels.



Although FairNAS \cite{chu2019fairnas} largely alleviates the ranking difficulty in the single-path case with a fairness strategy, it is inherently difficult to apply the same method in a multi-path scenario. It should be emphasized that the most critical point of the two-stage NAS is that the supernet can rank submodels as accurately as possible. However,  we find that a vanilla training of a multi-path supernet (e.g. randomly pick a multi-path submodel to train at each step) is not stable to provide a confident ranking because of changing statistics during the sampling process. Therefore, we dive into its real causes and undertake a unified approach to tackle this issue. Our contributions can be summarized as follows.


\begin{figure}[t]
	\centering
	\includegraphics[scale=0.43]{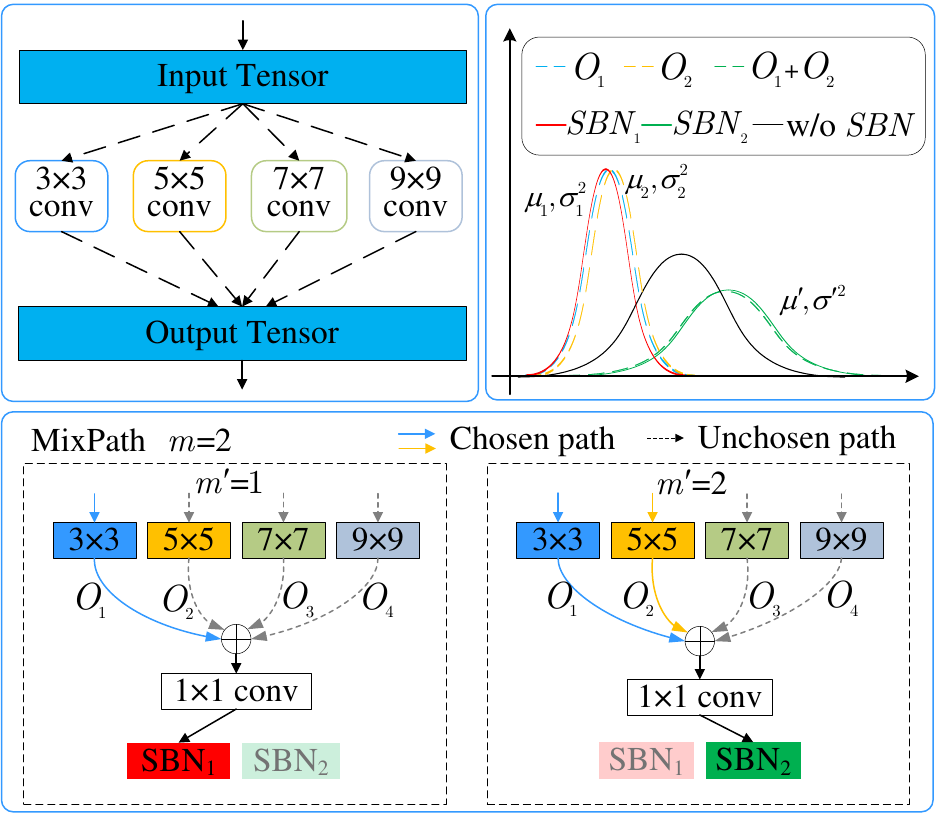}
	\caption{\textbf{Top Left}: Options in a demo block, where at most $m$ paths can be chosen. \textbf{Bottom}: An example of MixPath supernet training with Shadow Batch Normalizations (SBNs). Note 1x1 Conv is not a must. SBNs are standard BNs, except that SBNs are applied w.r.t the number of activated paths. 
		E.g., $SBN_1$ is used whenever only $m'=1$ path is activated, $SBN_2$ for $m'=2$ paths. \textbf{Top Right}: SBNs catch each of the statistics in two modes (red for $m'=1$, green for $m'=2$), however, a single BN (black) can't capture both.}
	\label{fig:mixpath-method}
\end{figure} 

%
\begin{itemize}
	\item We propose a unified approach for multi-path (say at most $m$ paths are allowed) one-shot NAS, as opposed to the current single-path methods. From this perspective, current one-shot methods \cite{guo2019single,chu2019fairnas} become one of our special cases when $m=1$.
	\item We disclose why a vanilla multi-path training fails, and we propose a novel yet lightweight mechanism, called \emph{shadow batch normalization} (SBN, see Fig.~\ref{fig:mixpath-method}), to stabilize the supernet with a neglectable cost. By exploiting feature similarity from different paths, we further reduce the number of SBNs to be $m$, otherwise it would be exponential.
	\item We reveal that our multi-path supernet is trained with stability due to SBNs. It also comes with boosted ranking performance. Together with post BN calibration, it obtains a high Kendall tau (0.597) on a subset of NAS-Bench-101 \cite{ying2019bench}.
	\item We use extensive experiments to verify the validness of our method across 4  spaces. We search proxylessly on ImageNet using10 GPU days. The searched models obtain state-of-the-art performances, which are comparable with MixNet \cite{tan2020mixconv} models searched with 200$\times$ more computing powers. Moreover, our MixPath-B makes use of multi-branch feature aggregation and reaches higher accuracy than EfficientNet-B0 with fewer FLOPS and parameters.
\end{itemize}

\section{Related work}
\label{gen_inst}

\textbf{Conditional batch normalization\quad} Batch Normalization (BN) \cite{ioffe2015batch} has greatly facilitated the training of neural networks by normalizing layer inputs to have fixed means and variances. 
Slimmable neural networks\cite{yu2018slimmable} introduces a shared supernet that can run with switchable channels at four different scales (1$\times$, 0.75$\times$, 0.5$\times$, 0.25$\times$). 
Due to feature inconsistency, independent BNs are applied for each switch configuration to encode conditional statistics. 
However, it is impractical because it requires an increased number of BNs when it has arbitrary channel widths. BatchQuant \cite{bai2021batchquant} utilizes BNs to stabilize the mixed-precision supernet training in a single-path MobileNetV3 search space. 

\textbf{Model ranking correlation\quad} It should be emphasized that the ranking ability for the family of two-stage algorithms is of the uttermost importance \cite{bender2018understanding}, whose sole purpose is to evaluate networks.  To quantitatively analyze their ranking ability, previous works like  \cite{Yu2020Evaluating,chu2019fairnas, li2019sgas, zheng2019multinomial} have applied a Kendall tau measure \cite{kendall1938new}.
To this end, various proxies \cite{tan2018mnasnet,zoph2016neural,zoph2017learning}, explicit or implicit performance predictors \cite{luo2018neural, liu2018progressive} are developed to avoid the intractable evaluation. Recent one-shot approaches \cite{bender2018understanding,guo2019single,chu2019fairnas} utilize a supernet where each submodel can be rapidly assessed with inherited weights. One-Shot \cite{bender2018understanding} uses dropout (a very sensitive hyper-parameter)to simulate sampling and evaluating submodel using a supernet. However, its poor ranking performance prevents it from  robustly finding good models \cite{chu2019fairnas}.

\section{MixPath: a unified approach}
\label{headings}

\subsection{Motivation}\label{sec:motivation}

\textbf{There is a call for an efficient multi-path one-shot approach.} Informally, existing weight-sharing approaches \cite{liu2018darts,chu2019fair,guo2019single,chu2019fairnas} can be classified into four categories based on two dimensions: stage levels and multi-path support, as shown in Fig.~\ref{fig:taxonomy}. Specifically, DARTS \cite{liu2018darts} and Fair DARTS \cite{chu2019fair} both are one-stage methods while the latter allows multiple paths between any two nodes. One-shot methods \cite{guo2019single,chu2019fairnas} typically train a supernet in the first stage and evaluate submodels in the second stage for model selection. So far, the two-stage one-shot methods mainly consider a single-path search space except \cite{bender2018understanding} which involves strong regularization and careful hyper-parameter tuning for supernet training. It is thus natural to study this multi-path counterpart further. MPNAS \cite{wang2021multi} resorts to reinforcement learning as in \cite{pham2018efficient} for a multi-path searching but verified only on small datasets. ScaleNAS \cite{cheng2022scalenas} adopts an evolutionary algorithm in a multi-scale search space but it is designed for downstream tasks like segmentations and human pose estimation.

Moreover, it's reasonable to design such a search space regarding two factors. \textbf{First}, multi-path feature aggregation is proven to be useful, such as Inception series \cite{szegedy2015going,ioffe2015batch,szegedy2016rethinking,szegedy2017inception}  and ResNeXt \cite{xie2017aggregated}. \textbf{Second},  it has the potential to balance the trade-off between performance and cost.

\begin{figure}[ht]
	\centering
	\begin{tikzpicture}[auto,node distance=0.8cm,
	semithick,scale=0.7, every node/.style={scale=0.7}]
	\tikzstyle{cell} = [rectangle, minimum width=1cm, minimum height=0.2cm,text centered, draw=black, fill=xablue]
	\tikzstyle{block} = [cell, fill=xablue,draw=black]
	\tikzstyle{feature} = [cell,minimum width=1cm, minimum height=1cm, fill=black!20!white]
	\draw [-latex] (-6,1) -- (3.5,1);
	\draw [-latex] (0,-2) -- (0,3.5);
	\node[cell] (xtl) at (-5.8, 2) {two-stage};
	\node[cell] (xtr) at (-5.8, 0) {one-stage};
	\node[cell,fill=xagreen] (ytt) at (-2.5, 3.3) {single-path};
	\node[cell,fill=xagreen] (ytb) at (2.5, 3.3) {multi-path};
	\node (a) at (-2.5,0.5) {DARTS \cite{liu2018darts}};
	\node (a0) at (-2.5,0) {ProxylessNAS\cite{cai2018proxylessnas}};
	\node (a1) at (-2.5,-0.5) {PDARTS \cite{chen2019progressive}};
	\node (a2) at (-2.5,-1) {PCDARTS \cite{xu2020pcdarts}};
	\node (a3) at (-2.5,-1.5) {AtomNAS \cite{mei2020atomnas}};
	\node [xshift=1cm] (b) at (1.5,-0.5) {Fair DARTS \cite{chu2019fair}};
	\node (c) at (-2.5,2) {FairNAS \cite{chu2019fairnas}};
	\node (e) at (-2.5,1.5) {SPOS \cite{guo2019single}};
	\node (f) at (-2.5,2.5) {DNA \cite{li2019blockwisely}};
	\node [xshift=1cm] (h) at (1.5,2) {MixPath};
	\draw [dashed, -latex] (b)+(0,0.2) to (h);
	\draw [dashed, -latex] (a1) to (b);
	\draw [dashed, -latex] (c) to (h);
        \node [xshift=1cm] (h) at (1.5,2.5) {One-Shot \cite{bender2018understanding}};
	\end{tikzpicture}
	\caption{A brief taxonomy of weight-sharing NAS. MixPath fills the empty space of two-stage multiple-path methods.}
	\label{fig:taxonomy}
\end{figure} 

Without loss of generality, say an inverted bottleneck has an input feature of $C_{in} \times H\times W$, whose number of intermediate and output channels are $C_{mid}$ and  $C_{out}$ (same as $C_{in}$) respectively. Its computational cost can be formulated as $c_{total} = 2HWC_{in}C_{mid}+k^2HWC_{mid}= 2HWC_{mid}(C_{in}+\frac{k^2}{2})$, where $k$ is the kernel size of the depthwise convolution. Usually, the value of $k$ is set to 3 or 5. Since $C_{in}$ typically dominates $\frac{k^2}{2}$, we can boost the representative power of depthwise transformation by mixing more kernels with a neglectable increased cost. This design can be regarded as a straightforward  combination of MixConv \cite{tan2020mixconv} and ResNeXt \cite{xie2017aggregated}.

\begin{figure}[ht]
	\begin{subfigure}{\columnwidth}
		\centering
		\includegraphics[width=0.85\linewidth,scale=0.55]{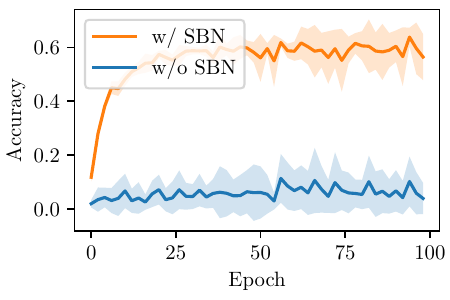}
		\caption{}
	\end{subfigure}
	\begin{subfigure}{\columnwidth}
		\includegraphics[width=\linewidth,scale=0.55]{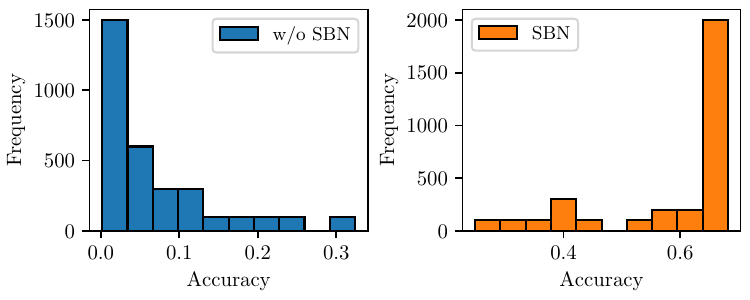}
		\caption{}
	\end{subfigure}
	\caption{\textbf{(a)} Average one-shot  validation accuracies (solid line) and their variances (shade) when training the MixPath supernet in $S_2$ (activating at most $m=2$ paths in each MixPath block) on ImageNet. Twenty models are randomly sampled every two epochs.  \textbf{(b)} Histogram of accuracies of randomly sampled 3k one-shot models. One-shot models generally come with good performance when SBNs are enabled during supernet training.}
	\label{fig:supernet_shadow_bn_comparison}
\end{figure}

\textbf{Multi-path supernet is very hard to train.\quad}Vanilla training of the multi-path supernet suffers from severe training difficulties. We can simply train a multi-path supernet by randomly activating a multi-path model at a single step. Here we randomly activate or deactivate each operation. However, this training process is very unstable according to our pilot experiments conducted with the MixPath supernet on ImageNet \cite{deng2009imagenet}, shown by the blue line in  Fig.~\ref{fig:supernet_shadow_bn_comparison} (a). 

\begin{figure*}[ht]
	\centering
	 \begin{subfigure}{0.55\linewidth}
		\includegraphics[width=\textwidth]{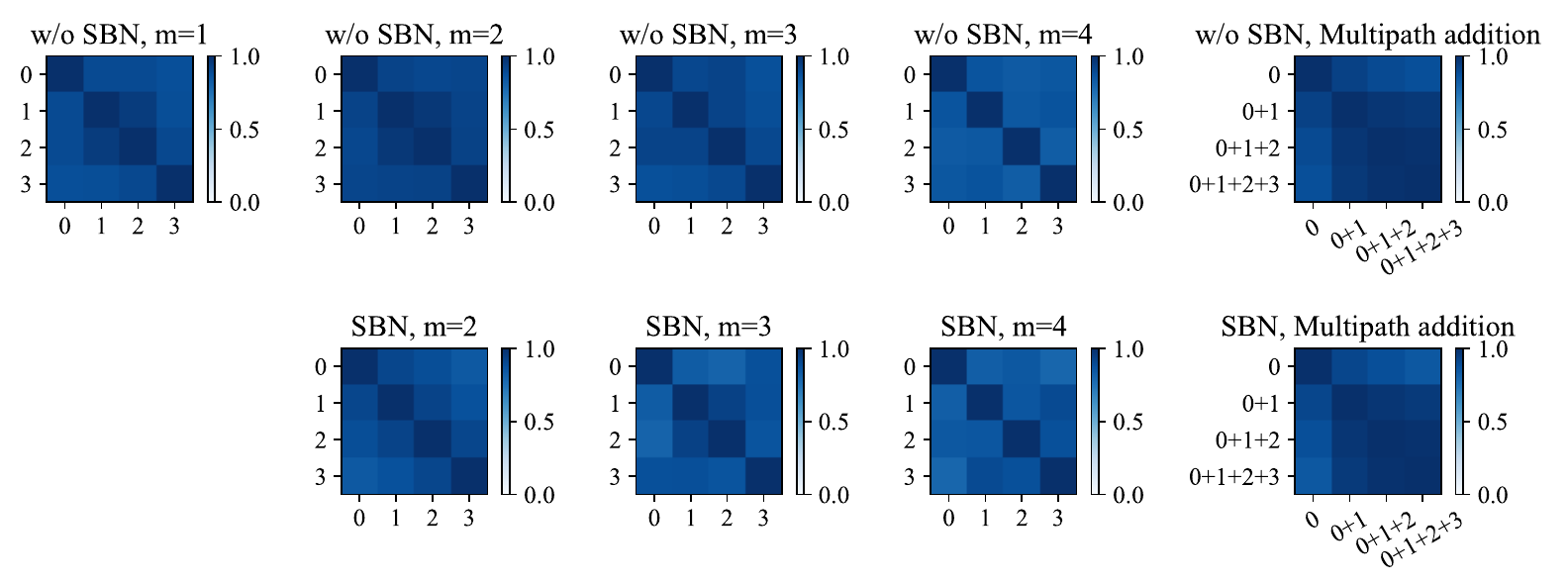}
		\caption{Similarity of multi-path feature vectors}
	\end{subfigure}
	\begin{subfigure}{0.35\textwidth}
		\includegraphics[width=0.45\columnwidth]{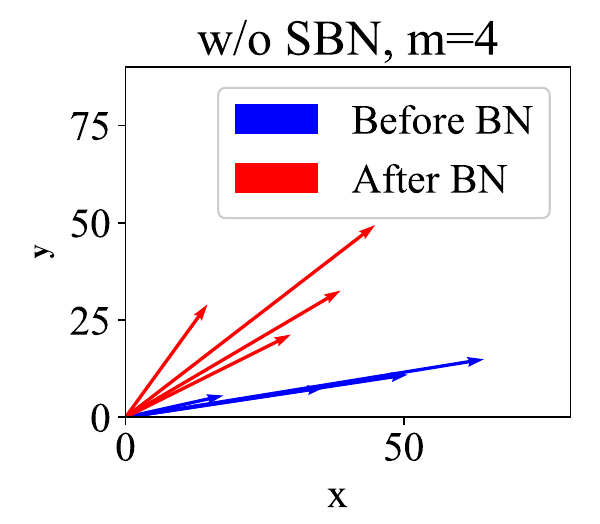}
		\includegraphics[width=0.45\columnwidth]{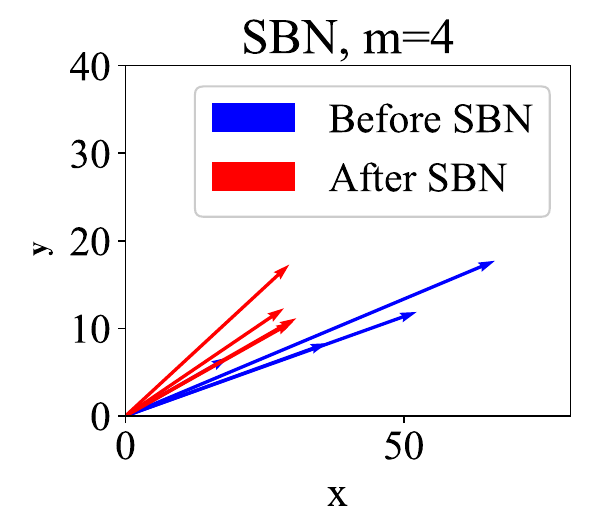}
		\caption{Feature vectors projected to 2d space}
	\end{subfigure}
	\caption{\textbf{(a)} Features from various path combinations are quite alike (w/ and w/o SBNs). Shown here is the cosine similarity map of first-block feature vectors from the supernets (in $S_1$). Darker means higher similarity. \textbf{(b)} SBNs regularize the magnitudes of feature vectors (projected into 2-dimensional space here) while vanilla BN cannot. Blue arrows represent the feature vectors obtained before SBNs or a single BN, and red ones after them. Here SBNs transform the magnitudes (19.6, 37.2, 53.7, 68.7) to around 32. BN normalized magnitudes (32.6, 36.6, 50.6, 66.7) are still diverse.} 
	\label{fig:cosine}
\end{figure*}

What might have caused such a problem? It has been  pointed out that high cosine similarity of features from different operations is important for training stability \cite{chu2019fairnas}. To verify the issue, we draw the cosine similarity matrix of different features from our trained supernet in the top row of Fig.~\ref{fig:cosine} (a). Surprisingly, not only are the features from single-path similar to each other, but also the multi-path features where they get added up. We discover that cosine similarity is limited because it measures the orientation only. The addition of high-dimensional feature vectors does not change their angles too much (as they are similar), but their magnitudes get scaled up, see blue arrows in Fig.\ref{fig:cosine} (b). We can further make a bold postulation that \emph{a simple superposition of multiple vectors renders very dynamic statistics, hence it causes training instability.} With this motivation in mind, we next scale the feature vectors down to the same magnitude to verify if we can make it stabilized. 

\subsection{Regularizing Multi-path Supernet}\label{sec:bn}

First of all, we can think of using a large amount of BNs to track the changing statistics. For a thorough regularization, we adopt multiple standard BNs to regularize feature statistics from different combinations of $m$ paths. We call them \emph{shadow batch normalizations} (SBNs) since they follow the activated paths combinations as shadows. Strictly speaking, in order to regularize all feature combinations in a multi-path setting, the number of SBNs has to grow exponentially.\footnote{$C_m^1 + C_m^2 + \cdots + C_m^m = 2^m-1$} 

\subsubsection{Reduce the complexity from $2^m$ to $m$.}
The above regularization is too costly. As \cite{chu2019fairnas} indicates for the single path case, the outputs of different operations at the same layer have higher similarities. It motivates us to reduce the number of shadow BN by diving into the underlying mechanisms further. 
Informally, if the features of every single path are similar (both magnitude and angles), we can derive that their combinations have similar statistics too, then the number of BNs can be reduced to $m$. Formally, this can be   formulated as follows.

\begin{definition}\label{def:zero-order}
	
		\textbf{Zero-order condition}: Let two high-dimensional random variables $\bf y=f(\bf x) \in R^{M\times d_1}$ and  $\bf z=g(\bf x) \in R^{M\times d_1}$,  we say that ${\bf y}$ and ${\bf z}$ satisfy the zero-order condition if  $|\bf y - \bf z |_2 \leq \epsilon$  for any valid sample $\bf x \in R^{N\times d}$, where $\epsilon$ is a very small positive float number. 
\end{definition}





\begin{lemma}\label{lem:m-comb}
	Let $\mathbf{m}$ be the maximum number of the activable paths, and each pair of operations satisfy the zero-order condition, we can use  $\mathbf{m}$ kinds of expectation and variance to approximate  all combinations (2$^m$).
\end{lemma}

The proof is provided in \ref{pr:lem-m-comb} (appendix). Note that the magnitude of $\epsilon$  largely affects the approximation error.  It is pretty straightforward as the expectations and variances are simply scaled up according to the number of additions. For instance, $\mathbb E[{\bf y} + {\bf z}] \approx \mathbb E[{\bf u} + {\bf v}] \approx 2  {\mathbb E} [{\bf y}]$. Lemma \ref{lem:m-comb} assures us that we can take $SBN_{i}(i=1, 2, \cdots, m)$ to track the combination that contains $i$ paths.
%
For the multi-path setting, we can observe similar results in Figure~\ref{fig:cosine}, which means the requirement of Definition~\ref{def:zero-order} is also satisfied. 



\begin{figure*}[t]
	\centering
	\includegraphics[scale=0.6]{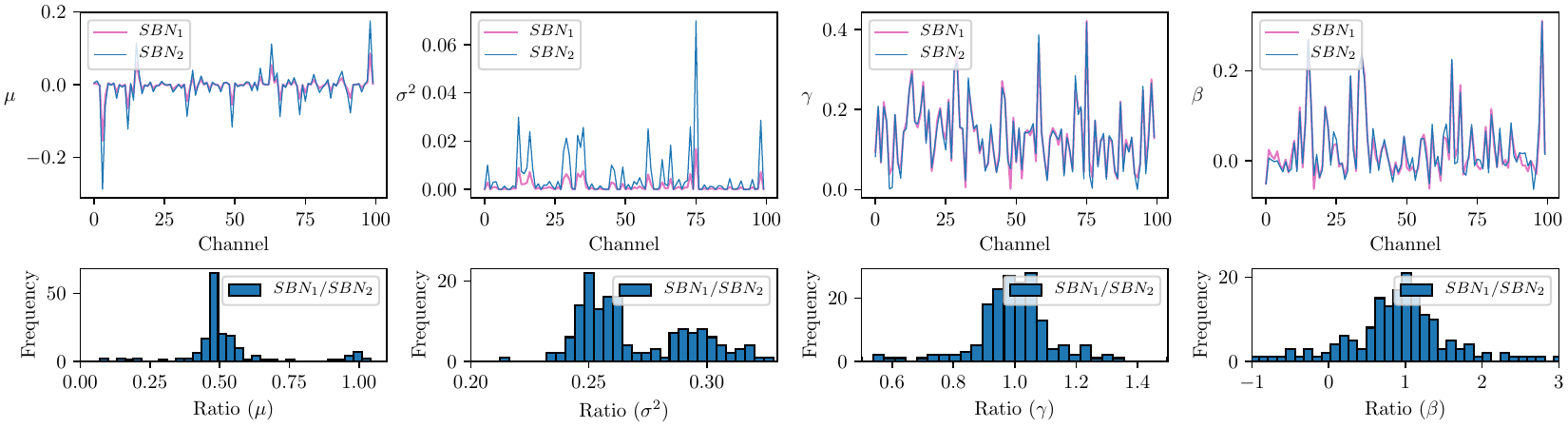}
	\caption{\textbf{Top}: Parameters ($\mu, \sigma^2, \gamma, \beta$) of the first-layer SBNs in MixPath supernet trained on CIFAR-10 when at most $m=2$ paths are activated. Specifically, $SBN_1$ and  $SBN_2$ represent SBNs for a single path and two paths. \textbf{Bottom}:  Histogram of ratios ($SBN_1/SBN_2$) for each BN parameter. They center around (0.5, 0.25, 1, 1). Note the mean of $SBN_2$ is roughly twice as that of $SBN_1$. It's interesting that we have  previously forecasted this result.}
	\label{fig:m2_shadow_bn_with_ratio}
\end{figure*}

\textbf{Statistical analysis of Shadow Batch Normalization.} We look into the statistics from the supernet trained on CIFAR-10\footnote{It's cheaper to train the supernet on CIFAR-10, and similar observations are also reproducible on ImageNet.} in search space $S_1$ (elaborated later in Section~\ref{sec: cifar10_s1}) (appendix).  Without loss of generality, we set $m$ = 2 and collect statistics of the four parameters of SBNs across all channels for the first choice block in Fig.~\ref{fig:m2_shadow_bn_with_ratio}. Note $SBN_1$ captures the statistics of one path, so does $SBN_2$ for two paths. Based on the theoretical analysis of Section~\ref{sec:bn}, we should have $\mu_{bn_1} \approx 0.5 \mu_{bn_2}$ and $\sigma_{bn_1}^2 \approx 0.25\sigma_{bn_2}^2$. This is consistently observed in the first two columns of Fig.~\ref{fig:m2_shadow_bn_with_ratio}. It's also interesting to see that the other two learnable parameters $\beta$ and $\gamma$ are quite similar for $SBN_1$ and $SBN_2$. This is explainable as SBNs have transformed different statistics to a similar distribution, the parameters of $\beta$ and $\gamma$ become very close.  Similar results are observed when $m = 3,4$  and in different layers in Fig.~\ref{fig:sbn-relation-m3-m4} and Fig.~\ref{fig:sbn-relation-m3-l6-l11} in \ref{subsec:bn-m-geq3} (appendix). 

\subsection{MixPath Neural Architecture Search}

Based on the above analysis, we train our multi-path supernet with \emph{shadow batch normalizations}, which we call the MixPath supernet. Following One-Shot \cite{bender2018understanding}, the supernet is used to evaluate models in the predefined search space, which makes our approach a two-stage pipeline: 1) training supernet with SBNs and 2) searching for competitive models. Specifically, in the training stage, we apply random sampling so that multiple paths can be activated at each step, where they add up to a mixed output. We then attach a corresponding SBN to the current path combination. This process is detailed in Algorithm \ref{alg:mixpath-super} (appendix). 
Next, similar to \cite{chu2019fair}, we progress with a well-known evolutionary algorithm NSGA-II \cite{deb2002fast} for searching. In particular, our objectives are to maximize classification accuracy while minimizing the FLOPS. In addition, batch normalization calibration \cite{bender2018understanding,guo2019single,mei2020atomnas}, as an orthogonal post-processing trick, is verified effective to further improve the ranking ability.

\begin{algorithm}[th]
	\caption{\textbf{: Stage 1} Supernet training with SBN.}
	\label{alg:mixpath-super}
	\begin{algorithmic}
		\STATE {\bfseries Input:} search space $S_{(n,L)}$ with $n$ choice paths per layer and $L$ layers in total, maximum optional paths $m$, supernet parameters $\Theta{(n,L)}$, training epochs $N$, training data loader $D$, loss function $Loss$.
		\STATE Initialize  every $\theta_{j,l}$ in  $\Theta_{(m,L)}$, SBN set $\{SBN_{1}, SBN_{2}, \cdots, SBN_{n}\}$ in each layer.
		\FOR{$i=1$ {\bfseries to} $N$}
		\FOR{$data,labels$ {\bfseries in} $D$}
		\FOR{$l=1$ {\bfseries to} $L$}
		\STATE {Randomly sample a number $m^{\prime}$ between 1 and $m$, then randomly sample $m^{\prime}$ paths without replacement, where each has output $\mathbf{o}_i^l$}, and use $SBN_{m^{\prime}}$ to act on the sum of outputs: $ \mathbf{o}^l =SBN_{m^{\prime}}\left(\sum_i \mathbf{o}_i^l \right) $
		\ENDFOR
		\STATE Select $model$ from above sampled index.
		\STATE  Clear gradients recorder for all parameters.
		\STATE Calculate gradients for $model$ based on $Loss$, $data$, $labels$, update $\theta_{(m,L)}$ by  gradients.
		\ENDFOR
		\ENDFOR
	\end{algorithmic}
\end{algorithm}

\section{Experiments}
\subsection{Search spaces}
We investigate four search spaces in this paper, from $S_{1}$ to $S_{4}$. For searching on CIFAR-10, we use $S_{1}$ that contains 12 inverted bottleneck blocks. For ImageNet, we search in $S_{2}$ and $S_{3}$ that are similar to MnasNet \cite{tan2018mnasnet} and MixNet \cite{tan2020mixconv}. For Kendall Tau analysis, we devise a subset of a common benchmark NAS-Bench-101 \cite{ying2019bench},  denoted as $S_{4}$. The details of these search spaces are listed in Table~\ref{table:ss} (appendix).

\subsection{Ranking ability analysis on NAS-Bench-101}\label{sec:nas101-experiment}

\begin{figure}[ht]
	\centering
	 \begin{subfigure}{\linewidth}\centering
		\includegraphics[width=0.8\textwidth,scale=0.5]{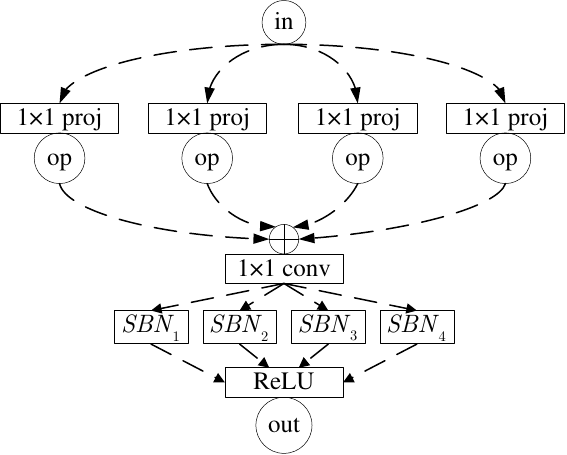}
	\end{subfigure}
 	\begin{subfigure}{\linewidth}\centering
		\includegraphics[width=0.9\textwidth,scale=1.0]{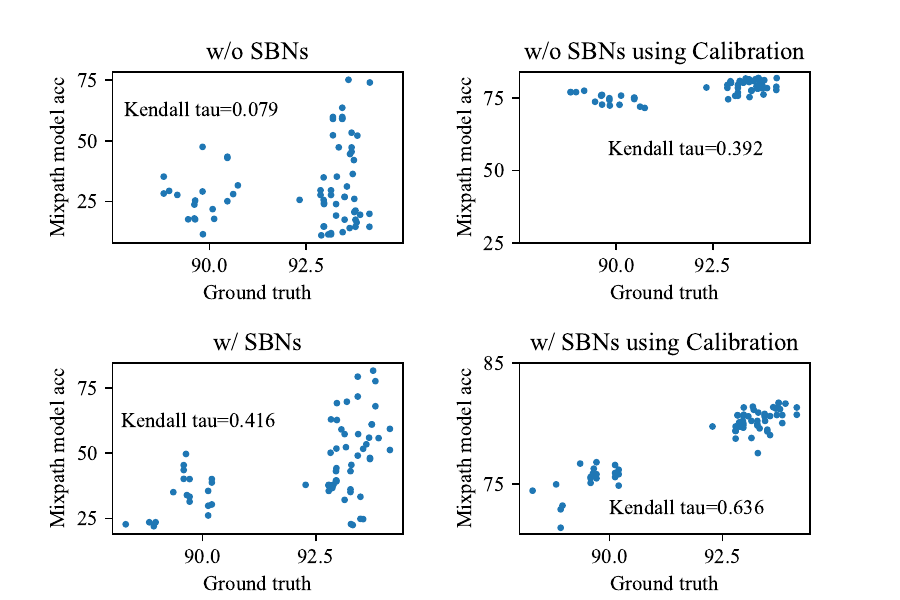}
	\end{subfigure}
	\caption{\textbf{(a)} Cell space with SBNs in NAS-Bench-101 \cite{ying2019bench} \textbf{(b)} Ground truth accuracies vs. predictions by one-shot MixPath supernets ($m=4$) trained with various strategies. We sample 70 models in NAS-Bench-101 \cite{ying2019bench} (sampling one time) to get the ground-truth. The supernet trained SBN with post BN calibration (top right) gives the best prediction.  A vanilla BN alone (w/ SBNs) wrongly pushes models into a narrow range under 0.2.}
	\label{fig:nasbench}
\end{figure}

To show that SBNs can stabilize the supernet training and improve its ranking, we score our method on a subset of a NAS-Bench-101 \cite{ying2019bench}. Each model is stacked by 9 cells, and every cell has at most 5 internal nodes. We make some adaptations to our method. The first 4 nodes are used to make candidate paths, each has 3 possible operations: $1 \times 1$ Conv, $3 \times 3$ Conv, and $3 \times 3$ Maxpool. The outputs of selected paths are summed up to give input to the fifth node (assigned as $1 \times 1$ Conv), after which the proposed SBNs are used. The designed cell space is shown in Fig.~\ref{fig:nasbench}.

Firstly, we conduct the baseline experiments on NAS-Bench-101 \cite{ying2019bench}, training with and without SBNs. For the latter, the features from the multi-path are summed up and then divided by the number of activated paths, following a $1 \times 1$ Conv-BN-ReLU (i.e., a single BN). 

To deeply analyze the model ranking capacity of our method and to understand the effects of SBNs, we conduct two pairs of experiments: 
\begin{enumerate*}[(i)]
	\item vanilla BN vs. SBN
	\item post-calibrated vanilla BN vs. post-calibrated SBN. 
\end{enumerate*}
For all experiments, we train the supernet for 100 epochs with a batch size of 96 and a learning rate of 0.025 on three different seeds. We randomly sample 70 models to find their ground-truth top-1 accuracy from NAS-Bench-101 to obtain Kendall tau \cite{kendall1938new}. Compared with FairNAS \cite{chu2019fairnas} and SGAS \cite{li2019sgas} utilizing about 10-20 models to calculate Kendall-tau, 70 models are adequate enough to evaluate the ranking performance.
 We use $m=4$ in this section.\footnote{We place $m=3$ case in ablation study.}
The results are shown in Table~\ref{tab:bn-calib-tau} and Fig.~\ref{fig:nasbench} (b).

\begin{table}[ht]
	\centering
	\caption{Comparison of Kendall taus between MixPath supernets ($m=4$) trained with various strategies. For each control group, we sampled 70 models from NAS-Bench-101 \cite{ying2019bench} and repeated it 3 times on different seeds. $^\dagger$: after BN calibration.}
	\label{tab:bn-calib-tau}

	\begin{tabular}{|l|c|c|}
		\hline
		Strategy & w/o Calibration & w/ Calibration  \\
		\hline\hline
		w/o SBN &  0.117$\pm$0.027 & 0.382$\pm$0.007 \\
		w/ SBN  & 0.393$\pm$0.037 &  \textbf{0.597$\pm$0.037} \\
		\hline
	\end{tabular}
	\end{table}

Notably, BN post-calibration can boost Kendall tau in each case, which confirms the validity of this post-processing trick \cite{bender2018understanding,guo2019single,mei2020atomnas}. Compared with baseline experiments, the supernet trained with SBNs has a much stronger ranking capacity. Even without calibration, the proposed method still ranks architectures better than the calibrated vanilla BN (+0.011). The composition of two tricks can further boost the final score to 0.597. Next, we will search for models in the search space commonly used for CIFAR-10 \cite{krizhevsky2009learning} and ImageNet \cite{deng2009imagenet} dataset respectively.

\subsection{Search on CIFAR-10}\label{sec: cifar10_s1}
We conduct experiments on CIFAR-10 dataset \cite{krizhevsky2009learning} with a designed search space $S_1$ containing 12 inverted bottlenecks \cite{sandler2018mobilenetv2}, each of which has four choices of kernel size (3, 5, 7, 9) for the depthwise layer and two choices (3, 6) for expansion rates. Hence, the size of search space $S_{1}$ ranges from $8^{12}$ ($m$=1) to $162^{12}$ ($m$=4). For each case, we train the supernet on CIFAR-10 for 200 epochs. The batch size is set to 96 and we use the SGD optimizer with a momentum of 0.9 and $3 \times 10^{-4}$ weight decay. We also use a cosine scheduler for the learning rate strategy with an initial value of 0.025. It only takes about 6 GPU hours on a single Tesla V100 GPU in total. The comparison with recent state-of-the-art models is listed in Table~\ref{tab:comparison-cifar}. Our searched model MixPath-c obtains 97.4 \% top-1 accuracy using 493M FLOPS on CIFAR-10, whose architecture is shown in Fig.~\ref{fig:mpos-architecture} in \ref{subsec:architecture} (appendix).

\begin{table}[ht]
	\small
	\centering
	\setlength{\tabcolsep}{2pt}
	\caption{Comparison on CIFAR-10. $^\dagger$: MultAdds computed from the provided code. $^\star$: transferred from ImageNet (search cost is on ImageNet). RL: Reinforcement learning, GD: Gradient-based, OS: one-shot, TF: Transferred.}
	\begin{tabular}{|l|r|r|r|r|r|} 
			\hline
			Models  & Params & $\times+$ & Top-1  & Type & Cost  \\
			&(M)&(M)&($\%$)& & {\small G $\cdot$ d} \\
			\hline\hline
		    NASNet-A \cite{zoph2017learning}  & 3.3 & 608  &  97.4 & RL & 1800 \\
		    NASNet-A Large \cite{zoph2017learning}$^\star$   & 85.0 & 12030 & 98.0 & TF & 1800 \\
		    ENAS \cite{pham2018efficient} & 4.6 & 626$^\dagger$ & 97.1 & RL   & 0.5  \\	
			DARTS \cite{liu2018darts} & 3.3 & 528 & 97.1 & GD & 0.4\\ 
			RDARTS \cite{zela2020understanding} &-& -& 97.1 & GD & 1.6 \\
			SNAS \cite{xie2018snas}  & 2.9 & 422 & 97.0 & GD & 1.5 \\
			GDAS \cite{dong2019searching} & 3.4 & 519 & 97.1 &GD &0.2 \\
			P-DARTS \cite{chen2019progressive} & 3.4 & 532 & 97.5 & GD &0.3\\ 
			PC-DARTS \cite{xu2020pcdarts} & 3.6 & 558 & 97.4 & GD & 0.1\\ 
			DARTS- \cite{chu2021darts}& 3.5& 568&97.5&GD&0.4\\
			Few-shot DARTS \cite{zhao2021few} & 3.8 & - & 97.6 & GD & 1.3 \\
			MixPath-c (ours) & 5.4 & 493 & 97.4 & OS & 0.25 \\
			\hline
			MixNet-M \cite{tan2020mixconv}$^\star$ & 3.5  & 360 & 97.9 & TF & $\approx$3k \\
			MixPath-a (ours)$^\star$ & 3.5 &348 & 98.1 &TF & 10\\
			\textbf{MixPath-b} (ours)$^\star$ & 3.6  & 377 & 98.2 & TF & 10 \\

			\hline
		\end{tabular}
		\label{tab:comparison-cifar}
\end{table}

\subsection{Proxyless Search on ImageNet }
We search proxylessly on ImageNet \cite{deng2009imagenet} in the search space of MnasNet \cite{tan2018mnasnet}, as this layer-wise search space with inverted bottleneck blocks \cite{sandler2018mobilenetv2} is also very commonly used \cite{tan2018mnasnet,tan2019efficientnet,wu2018fbnet,chu2019fairnas,tan2020mixconv}. We fix the expansion rate as in \cite{tan2020mixconv} and search for the combinations of various depthwise convolutions.   
Specifically, we search from the kernel sizes (3, 5, 7, 9) and their combinations. We call this search space $S_2$. Moreover, we also construct a search space $S_3$ to incorporate architectures like MixNet \cite{tan2020mixconv}, where we evenly divide the channel dimension of the depthwise layer into 4 groups and search from the kernel sizes $(3, 5, 7, 9)$ for each group. We set $m=2$ (in $S_2$) and $m=1$ (in $S_3$). In each case, we utilize the same hyper-parameters: a batch size of 1024 and the SGD optimizer with 0.9 momentum and a weight decay of $10^{-5}$. The initial learning rate is 0.1 with cosine decay and it decreases to 0 within 120 epochs. This takes about 150k times of backpropagation (10 GPU days on Tesla V100).  Once the supernet is trained, we adopt NSGA-II \cite{deb2002fast} for model selection. 


As per training stand-alone models, we use the same tricks as MixNet-M \cite{tan2020mixconv}, the results are listed in Table~\ref{tab:comparison-imagenet}. To make fair comparisons, we disable distillation in NAS approaches because it can marginally boost the performance of found models. Our cost in Table~\ref{tab:comparison-imagenet} includes training supernet (10 GPU days) and NSGA-II (0.3 GPU day). MixPath-A, sampled from the Pareto front in $S_3$ (so is MixNet), uses 349M multiply-adds to obtain $76.9\%$ accuracy on ImageNet. MixPath-B, in $S_2$, uses fewer FLOPS and parameters w.r.t EfficientNet-B0 to obtain $77.2\%$. Note SPOS, FBNet, ProxylessNAS are also in $S_2$ or similar. In the appendix we list the \textbf{detailed structures} of two models in Fig.~\ref{fig:mpos-architecture} of \ref{subsec:architecture}. 

\paragraph{Transferability.} We also transfer these models to CIFAR-10 and \textit{COCO} detection, please refer to Table~\ref{tab:comparison-cifar} and \ref{subsec:transfer-coco}.


\begin{table}[ht]
	\small
	\centering
	 \caption{Comparison with state-of-the-art models on ImageNet. For fair comparisons, we disable distillations for all methods. Search cost is computed by GPU days. $^\dagger$: TPU days. $^\ddagger$: w/ distillation.}
	\label{tab:comparison-imagenet}
	\smallskip
	 \begin{small} 
		\setlength{\tabcolsep}{3pt}
		\begin{tabular}{|l|l|l|l|l|r H|} 
			\hline
			Models  & $\times$+  & \#P  & Top-1 & Top-5 & Cost & Latency\\
			&(M)&(M)&($\%$)&($\%$) & {\small G $\cdot$ d} & (ms)\\
			\hline\hline
			MobileNetV2 \cite{sandler2018mobilenetv2}  & 300  & 3.4 & 72.0 & 91.0 & - & 8.6$\pm$0.5\\
			ShuffleNetV2 \cite{ma2018shufflenet} & 299 & 3.5 & 72.6 & - & - & 10.6$\pm$0.6\\
			GhostNet 1.3$\times$ \cite{han2019ghostnet} & 226 & 7.3 & 75.7 & 92.7 & -& \\
			\hline
			DARTS \cite{liu2018darts} & 574 & 4.7 & 73.3 &-  & 0.5 & 29.4$\pm$1.1\\
			P-DARTS\cite{chen2019progressive}  & 557 & 4.9 &75.6 & 92.6 & 0.3 & 30.5$\pm$1.7\\
			PC-DARTS \cite{xu2020pcdarts} &586 &5.3& 74.9 & 92.2 & 0.1 & 33.5$\pm$0.4\\
			SGAS (Cri.2 best) \cite{li2019sgas} & 598 & 5.4 & 75.9 & 92.7 & 0.25 & 34.7$\pm$0.5\\
			FairDARTS-C \cite{chu2019fair} & 380 & 4.2 & 75.1 & 92.4 & 0.4 & -\\
			MnasNet-A2 \cite{tan2018mnasnet}  & 340 & 4.8 & 75.6 & 92.7 & $\approx$4k & -\\
			One-Shot Small (F=32) \cite{bender2018understanding} &- & 5.1 & 74.2 & -& 4$^\dagger$ & -\\
			\hline
			FBNet-B \cite{wu2018fbnet} & 295 & 4.5 & 74.1 & - & 9 & - \\ 
			MobileNetV3 \cite{howard2019searching} & 219 & 5.4  &75.2 & 92.2 & $\approx$3k & -\\
			Proxyless-R \cite{cai2018proxylessnas}   & 320 & 4.0 & 74.6 & 92.2 & 8.3 & 7.4$\pm$0.2\\
			FairNAS-A \cite{chu2019fairnas}  & 388& 4.6 &  75.3 & 92.4 & 12 & 6.9$\pm$0.2\\ 
			Single Path One Shot \cite{guo2019single}  & 328 & 3.4 &74.9 & 92.0 & 12 & 10.2$\pm$0.3\\
			Single-Path NAS \cite{stamoulis2019single} & 365 & 4.3 & 75.0 & 92.2 & 1.25 & -\\

            OFA w/o PS \cite{Cai2020Once-for-All:} & 235 &4.4& 72.4 &-&50&\\
            OFA $^\ddagger$n \cite{Cai2020Once-for-All:} & 230 &- & 76.0&-&50&\\
            BigNAS-M $^\ddagger$ \cite{yu2020bignas} & 418 & 5.5 &77.4 &93.5& -&\\
			MixNet-M \cite{tan2020mixconv} & 360 & 5.0 &77.0& 93.3 & $\approx$3k & 23.0$\pm$0.8 \\
			MixPath-A  (ours)& 349 & 5.0 & 76.9 & 93.4 & 10.3 &29.1$\pm$0.3\\
			MixPath-B (ours) & 378 & 5.1 & 77.2 & 93.5 & 10.3 & 20.9$\pm$0.4 \\
			\hline
		\end{tabular}
		\setlength{\tabcolsep}{1.4pt}
	\end{small} 
\end{table}

\section{Ablation Study and Discussions}


\subsection{Supernet ranking with different strategies}
To study the individual contribution of SBN and BN post-calibration, we profit from NAS-Bench-101 \cite{ying2019bench} for ranking calculation since it provides ground truths for all models. Moreover, it allows a fair comparison with other approaches in the same search space. Specifically, we investigate the case $m=3$ and report the result of each control group using 3 different seeds. Kendall tau \cite{kendall1938new} is calculated by randomly sampled 70 models.

\textbf{Supernet with and without SBNs.} Table~\ref{tab:bn-calib-vs-sbntau} shows that the supernet trained without SBN is inadequate to evaluate sampled models ($\tau=0.05$), which is much lower than the one with a linear number of SBNs ($\tau=0.32$). We further quantify the effect of the additional post BN calibration, and both taus can be boosted. Particularly for SBN, it's improved to 0.59 ($+0.27$). Therefore, the combination of SBN and post BN calibration delivers the best ranking performance.

\begin{table}[ht]
	\caption{Kendall taus' comparison between MixPath supernets ($m=3$) trained with multiple strategies. (same settings as Table~\ref{tab:bn-calib-tau}).  ESBN: Exponential SBNs, LSBN: Linear SBNs }
	\label{tab:bn-calib-vs-sbntau}
	\smallskip
	\centering
	\small
	\setlength{\tabcolsep}{3pt}
	\begin{tabular}{|l|c|c|c|c|}
		\hline
		$\tau$ & w/o SBN   & ESBN & LSBN  \\
		\hline\hline
		w/o Calibration & 0.05$\pm$0.06  &  0.37$\pm$0.06 & 0.32$\pm$0.03 \\
		w/ Calibration &  0.43$\pm$0.03  &   0.34$\pm$0.08  & \textbf{0.59$\pm$0.02}  \\
		\hline
	\end{tabular}
\end{table}

	
\textbf{Linear number of SBNs vs. exponential.} Through theoretical analysis, we conclude that a linear number of SBNs are enough to match all the combinations of activable paths, while we can still use its exponential counterpart where every single combination follows an SBN. Table~\ref{tab:bn-calib-vs-sbntau} shows that linear SBN with BN calibration works best (with the highest $\tau$), while Exp SBN (exponential) performs best if no calibration is applied. 
	Why is it so? Linear SBNs make use of the zero-order condition which is an approximate relationship. Exp SBNs are more accurate since they catch statistics from each of the combinations.  However, this makes a wider distribution of learnable parameters $\gamma, \beta$, which is difficult for BN calibration to fit them well. Therefore, we choose Linear SBNs with BN calibration instead. Regarding the Occam's Razor, the linear growing BN generalizes better than the exponential counterpart.



\begin{figure*}[tb!]
	\centering
	\includegraphics[scale=0.6]{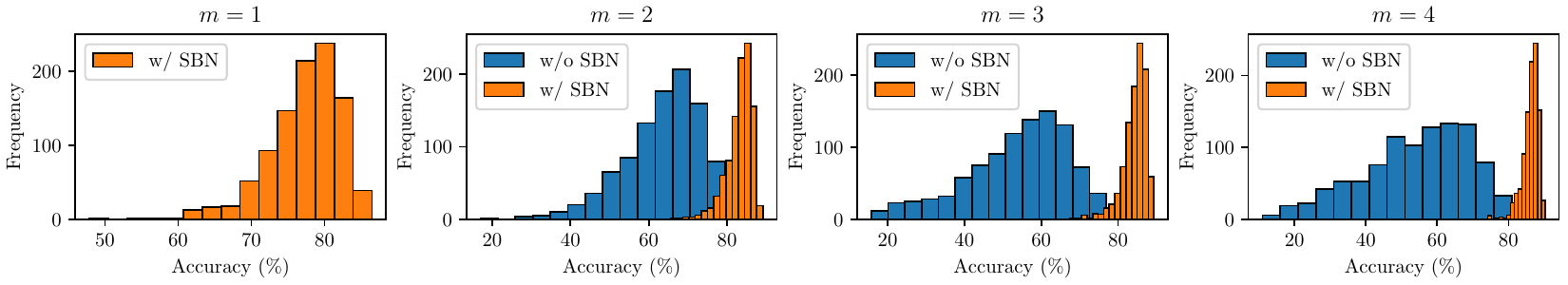}
	\caption{Histogram of one-shot model accuracies sampled from MixPath supernets trained on CIFAR-10 in $S_1$, activating at most $m=1,2,3,4$ paths respectively. In all cases, the supernet trained without SBNs severely underestimates a large proportion of candidate architectures. Note when $m=1$ it falls back to a vanilla BN.}
	\label{fig:multi_path_bn_comparison}
	\vskip -0.1in
\end{figure*}

\textbf{One-shot accuracy distribution of candidate models w.r.t. $m$}  We run four experiments with $m=1,2,3,4$ to investigate the effect of SBNs in $S_1$, other settings are kept the same. In total, 1k models are randomly sampled to plot their test accuracy distributions on CIFAR-10 for each $m$ in Fig.~\ref{fig:multi_path_bn_comparison}. When $m=1$, MixPath falls back to single path. Whereas, SBN begins to demonstrate its power for $m>1$, whose absence leads to a bad supernet with lower predicted accuracies and a much larger gap. This indicates the supernet without SBNs severely underestimates the performance of a large proportion of architectures. 

\textbf{Simple scaling vs. SBNs.}  Intuitively from the above motivation, simply scaling the magnitudes of all features to the same level in each block could also stabilize the supernet training. We hereby treat \emph{simple scaling} as a baseline experiment. However, it is somewhat limited because it adjusts magnitudes only.  We find that ranking ability after the calibration of the simple scaling case is higher than that of vanilla BNs and exponential SBNs, but it is lower than that of linear SBNs. This is consistent with the baseline experiment in Table~\ref{tab:sbn-ss}. Our proposed method with linear SBNs is still the best among all the compared strategies. Note that the simple scaling approach can only roughly scale the magnitudes of the output. In contrast, SBNs can adjust the magnitudes and angles together to better capture the changing distributions.

\begin{table}[ht]
	\caption{Comparison of Kendall taus between MixPath supernets ($m=4$) trained with various strategies. We sampled 70 models from NAS-Bench-101 \cite{ying2019bench} and repeated it 3 times on different seeds. $^\dagger$: after BN calibration}
	\label{tab:sbn-ss}
	\smallskip
	\centering
	\small
	\begin{tabular}{|l|c|c|}
		\hline
		Strategy & Simple Scaling & w/ SBNs  \\
		\hline\hline
		w/o Calibration &  0.293$\pm$0.016 & 0.393$\pm$0.017  \\
		Calibration  & 0.472$\pm$0.021 & \textbf{0.597$\pm$0.037} \\
		\hline
	\end{tabular}
\end{table}

\paragraph{Applying SPOS on multi-path search space.}\label{supp:spos-branch-comp}

\begin{figure}
	\centering
	\includegraphics[scale=0.67]{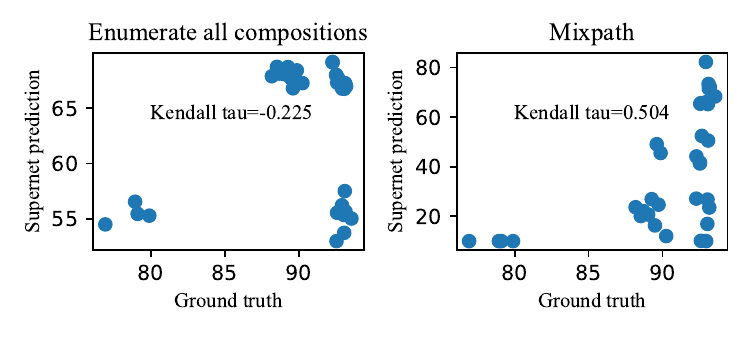}
	\caption{Rank performance comparison between `enumerating all compositions' and our method.}
	\label{fig:comp_mixpath_rank}
\end{figure}

We design a ``branch composition'' experiment $m=4$ in NAS-Bench-101 with 3 candidate operations, leading to 34 possible blocks. We use SPOS \cite{guo2019single} to train this supernet with the same hyper-parameters as ours, but we only get a Kendall tau of \textbf{-0.225} while our method obtains \textbf{0.504} in terms of ranking the same sub-models (see Fig.~\ref{fig:comp_mixpath_rank}). This result is not surprising. We blame the failure of the former approach for the use of an excessive number of blocks, which brings nearly \textbf{4$\times$} more trainable parameters in the supernet (199.7M vs. ours 52.9M). In this case, it's impractical and infeasible for SPOS to work under such setting because it's hard to make candidate blocks fully optimized. In effect, thus-trained supernet is unable to ensure a reliable rank. For example, on average, one block receives an update every 34 iterations, resulting in poor ranking performance. Besides, this approach will encounter memory explosion at a growing $m$. Even if we can solve the ranking issue, it requires much more GPU memory to save the trainable parameters, which prevents proxyless search on large datasets or big models. In contrast, our method doesn't suffer from these issues. 

\subsection{Comparison of search strategies}\label{sec:search-strategy}
We use the search space $S_3$ on ImageNet to compare the adopted NSGA-II search algorithm \cite{deb2002fast} with random search by charting the Pareto-front of models found by both methods in Fig.~\ref{fig:nsga-vs-random} (appendix). NSGA-II has an  advantage that the elite models have higher accuracies and fewer FLOPs.

\subsection{Pipeline component analysis}
Apart from the ablation studies on NAS-Bench-101 \cite{ying2019bench}, we make a component analysis of our pipeline in $S_1$. We focus on the supernet training with and without SBN by fixing the second-stage searching algorithm. Each pipeline is run three times on different random seeds. Our goal is to find the best models under 500M FLOPS. As a result, with SBN enabled, we obtain 97.44\% top-1 accuracy, higher than the best 97.12\% when SBN is disabled. This confirms that it mainly benefits from our novel training mechanism.

\subsection{Discussions}
\textbf{Why can SBNs stabilize supernet training and improve ranking ability?\quad} 
We have verified that high cosine similarity is not the only factor that guarantees the stable training of a multi-path supernet. It is more important to ensure that the feature distributions are consistent. Otherwise, the training is disrupted because a single BN is not enough to capture changing statistics from multiple paths. The proposed SBNs are able to track a variety of distributions, alleviating this instability. Empirically, better training of the supernet also gives more appropriate weights for each subnetwork, with which we can better estimate the ground-truth performance of the stand-alone models.
	
	
\textbf{Why can SBNs work together with BN post-calibration?\quad} According to the analysis in Section \ref{sec:bn}, the parameters $\mu, \sigma$ of SBNs that follow a multi-path combination are multiples of those of a single-path, as shown in  Fig.~\ref*{fig:m2_shadow_bn_with_ratio}. But not all the parameters fit well into this relationship, post BN calibration can then readjust the mean and variance of the inputs to fit the learned parameters $\gamma$ and $\beta$. To some extent, it can make up for the gap from the supernet to submodels. Noticeably, SBN is a type of regularization technique during the supernet training, while BN calibration is a post-processing trick. They can work in an orthogonal way and both contribute to the ranking.

\textbf{Comparison with Switchable Batch Normalization\quad}  Switchable BN \cite{yu2018slimmable} is only applied to learn the statistics of a limited number ($K$) of sub-architectures with $K=\{4,8\}$ channel configurations. Our SBN is designed to match the changing feature statistics from a more flexible combination of different choice paths, while the number of channels is fixed. The former is used for network pruning while our SBN is to stabilize the training of the multi-path supernet.

\section{Conclusion}
In this paper, we propose a unified approach for two-stage one-shot neural architecture search, with a multi-path search space enabled. Existing single-path approaches can thus be regarded as a special case. The proposed method uses SBNs to catch the changing features from various branch combinations, which successfully solves two difficulties of vanilla multi-path methods: the unstable training of supernet and the unbearable weakness of model ranking. Moreover, by exploiting feature similarities from different path combinations, we reduced the number of required exponential SBNs to $m$, identical to the number of allowed paths. Our method can enjoy the model ranking capacity of the one-shot supernet and robustly search good models across various benchmarks. 

\textbf{Acknowledgements}: This work was supported by National Key R\&D Program of China (No. 2022ZD0118700).

{\small
\bibliographystyle{ieee_fullname}
\bibliography{egbib}
}

\appendix

\section{Proofs}\label{sec:proof}

\subsection{Proof of Lemma 3.1}
\begin{proof}\label{pr:lem-m-comb}
	Let ${\bf y}_{p_{{\bf y}}({\bf y})}=f({\bf x})$, ${\bf z}_{p_{{\bf z}}({\bf z})} = g({\bf x})$, ${\bf x} \sim p_{{\bf x}}({\bf x})$. For the case $m=1$, the expectation of ${\bf y}$ and ${\bf z}$ can be written respectively as: 
	
	\begin{equation}
		\begin{aligned}
			{\mathbb E}[{\bf y}]&={\mathbb E}[f\left({\bf x}\right)]=\int p_{{\bf x}}({\bf x}) f({\bf x})  {\mathrm d} \, {\bf x} \\
			{\mathbb E}[{\bf z}]&={\mathbb E}[g\left({\bf x}\right)]=\int p_{{\bf x}}({\bf x}) g({\bf x})  {\mathrm d} \, {\bf x}
		\end{aligned}
	\end{equation}
	
	According to the zero-order condition, we have $f({\bf x}) \approx g({\bf x})$. And $p({\bf x})$ is same for both ${\bf y}$ and ${\bf z}$, so ${\mathbb E}[{\bf y}] \approx {\mathbb E}[{\bf z}]$. 
	
	Now we prove $Var[{\bf y}] \approx Var[{\bf z}]$. Note that $Var[{\bf y}] = {\mathbb E}\left[{\bf y}^{2}\right] - \left({\mathbb E}[{\bf y}] \right)^{2}$ and $Var[{\bf z}] = {\mathbb E}\left[{\bf z}^{2}\right] - \left({\mathbb E}[{\bf z}] \right)^{2}$, thus we only need to prove ${\mathbb E}\left[{\bf y}^{2}\right] \approx {\mathbb E}\left[{\bf z}^{2}\right]$. It can be  similarly proved as follows:
	
	\begin{equation}
		\begin{aligned}
			{\mathbb E}\left[{\bf y}^{2}\right]&=\int p_{{\bf y}}({\bf y}) {\bf y}^{2} {\mathrm d}  {\bf y} = \int p_{{\bf x}}({\bf x}) f^{2}({\bf x}) {\mathrm d} {\bf x}
			\\
			{\mathbb E}\left[{\bf z}^{2}\right]&=\int p_{{\bf z}}({\bf z}) {\bf z}^{2} {\mathrm d}  {\bf z} = \int p_{{\bf x}}({\bf x}) g^{2}({\bf x}) {\mathrm d} {\bf x}
		\end{aligned}
	\end{equation}
	
	According to the zero-order condition, we have $Var[{\bf y}] \approx Var[{\bf z}]$.
	
	For the case of $m=2$, when the two paths are both selected, the output becomes ${\bf y} + {\bf z}$, its expectation can be written as:
	
	\begin{equation}
		\begin{aligned}
			{\mathbb E}[{\bf y} + {\bf z}] &= {\mathbb E}[{\bf y}] + {\mathbb E}[{\bf z}] \approx 2{\mathbb E}[{\bf y}]
		\end{aligned}
	\end{equation}
	
	And the variance of ${\bf y} + {\bf z}$ is,
	
	\begin{equation}
		\begin{aligned}
			Var[{\bf y} + {\bf z}] \approx  Var [2{\bf y}] = 4 Var[\bf y]
		\end{aligned}
	\end{equation}
	
	Therefore, there are two kinds of expectations and variances: ${\mathbb E}[{\bf y}]$ and $Var[\bf y]$ for $\{{\bf y}, {\bf z}\}$, and $2{\mathbb E}[{\bf y}]$ and $4Var[\bf y]$ for $\{{\bf y}+{\bf z}\}$. Similarly, in the case where $m \in [1, n]$, there will be $m$ kinds of expectations and variances. 
\end{proof}


%


\section{Algorithms}

\begin{algorithm}[H]
	\caption{\textbf{: Stage 2}-NSGA-II search strategy.}
	\label{alg:search}
	\begin{algorithmic}
		\STATE {\bfseries Input:} Supernet $S$, the number of generations $N$, population size $n$, validation dataset $D$, constraints $C$, objective weights $w$.
		\STATE {\bfseries Output:} A set of $K$ individuals on the Pareto front. 
		\STATE Uniformly generate the populations $P_{0}$ and $Q_{0}$ until each has $n$ individuals satisfying $C_{acc}, C_{FLOPs}$.
		\FOR{$i=0$ {\bfseries to} $N-1$}
		\STATE $R_{i}=P_{i} \cup Q_{i}$
		\STATE $F=\text{non-dominated-sorting}(R_{i})$
		\STATE Pick $n$ individuals to form $P_{i+1}$ by ranks and the crowding distance weighted by $w$.
		\STATE $Q_{i+1} = \emptyset$
		\WHILE{$size(Q_{i+1})<n$}
		\STATE $M=\text{tournament-selection}(P_{i+1})$
		\STATE $q_{i+1}=\text{crossover}(M) $
		\IF{$FLOPs(q_{i+1})>FLOPs_{s_{max}}$}
		\STATE continue
		\ENDIF
		\STATE Evaluate model $q_{i+1}$ with $S$ (BN calibration is recommended) on $D$ 
		\IF{$acc(q_{i+1})>acc_{min}$}
		\STATE Add $q_{i+1}$ to $Q_{i+1}$
		\ENDIF
		\ENDWHILE
		\ENDFOR
		\STATE Select $K$ equispaced models near Pareto-front from $P_{N}$
	\end{algorithmic}
\end{algorithm}

\section{Experiments details}\label{sec:app-exp}

\subsection{Search Spaces}

We show the list of used search spaces in Table~\ref{table:ss}.

\begin{table*}[ht]
	\begin{center}
		\begin{tabular}{|c|c|c|m{2cm}|m{7cm}|}
			\hline
			Space & Dataset & $m$ & Size & Details \\
			\hline\hline
			$S_{1}$ & CIFAR-10  &1 & $8^{12}$ &  \multirow{4}{7cm}{There are 12 stacked inverted bottleneck blocks. Kernel sizes are in (3, 5, 7, 9), and expansion rates are in (3, 6).} \\
			& & 2 &$36^{12}$ &  \\
			& & 3 &$92^{12}$  & \\
			& & 4 &$162^{12}$ &  \\
			\hline
			$S_{2}$ & ImageNet  & 2 & $10^{18}$  &There are 18 stacked inverted bottleneck blocks. Kernel sizes are in (3, 5, 7, 9). Expansion rate is fixed following MixNet. \\
			\hline
			$S_{3}$ & ImageNet  & 1 & $256^{18}$  &There are 18 stacked mobile inverted bottlenecks. Depthwise layer channels are divided into 4 groups and there are 4 choice kernel sizes (3, 5, 7, 9) for each group. Expansion rate is also fixed as above.\\
			\hline
			$S_{4}$ & CIFAR-10  & 4 &$255$ &There are 9 stacked cells, each with 5 internal nodes, where the first 4 nodes are candidate paths. Each node has 3  operation choices ($1 \times 1$ Conv, $3 \times 3$ Conv, $3 \times 3$ Maxpool). See Fig.~\ref{fig:nasbench} (main text). \\
			\hline
		\end{tabular}
		\caption{Four search spaces used in this paper.  $m$ is the maximum number of allowed paths per layer}
		\label{table:ss}
	\end{center}
\end{table*}
\setlength{\tabcolsep}{1.4pt}

\subsection{More experiments }\label{sec:s4}

We further search directly in $S_4$. To be comparable, this case is formulated as a single objective optimization problem: finding the best model with known ground truth (94.29\%) in the space. As a strong baseline, we run DARTS \cite{liu2018darts} three times using different seeds. The result is shown in Table~\ref{tab:nas101_comparison}. Our method obtains $94.22\%$ within 5 GPU hours, which outperforms DARTS with a large margin.

\begin{table}[ht]
	\centering
	\caption{Search results on the reduced NAS-Bench-101. The accuracy of known optimal is 94.29\%.}
	\smallskip
	\label{tab:nas101_comparison}
	\begin{tabular}{|l|c|c|}
		\hline
		Method &Top-1 Acc & Search Cost \\
		&  (\%) & (GPU Hours) \\
		\hline\hline
		DARTS \cite{liu2018darts}  & 79.42$\pm$0.23 &  7\\
		Ours & \textbf{94.22$\pm$0.06 }&   5\\
		\hline
	\end{tabular}
\end{table}

We  also use $m=3$ and perform  multi-path supernet training in DARTS space. Then we search for the optimal sub-model with 60 generations. The total search cost is 0.5 GPU days and MixPath achieves a competitive 97.5\% test accuracy with only 3.6M parameters on CIFAR10.

\subsection{Transferring to CIFAR-10}

We also evaluated the transferability of MixPath models on CIFAR-10 dataset, as shown in Table \ref{tab:comparison-imagenet} (main text). The settings are the same as \cite{huang2018gpipe} and \cite{kornblith2019better}. Specifically, MixPath-b achieved $98.2\%$ top-1 accuracy with only 377M FLOPS.


\subsection{Transferring to object detection}\label{subsec:transfer-coco}
We further verify the transferability of our models on object detection tasks and we only consider mobile settings. Particularly, we utilize the RetinaNet framework \cite{lin2017focal}  and use our models as drop-in replacements for the backbone component. Feature Pyramid Network (FPN) is enabled for all experiments. The number of the FPN output channels is 256. The input features from the backbones to FPN are the output of the depth-wise layer of the last bottleneck block in four stages, which covers 2 to 5. 

All the models are trained and evaluated on the MS COCO dataset \cite{lin2014coco} (train2017 and val2017 respectively) for 12 epochs with a batch size of 16. We use the SGD optimizer with 0.9 momentum and 0.0001 weight decay.
The initial learning rate is 0.01 and multiplied by 0.1 at epochs 8 and 11. Moreover, we use the MMDetection toolbox \cite{chen2019mmdetection} based on PyTorch \cite{paszke2019pytorch}. Table~\ref{table:mixpath-coco-retina}  shows that MixPath-A gives competitive results.

\begin{table}
	\begin{center}
		\small
		\setlength{\tabcolsep}{1pt}
		\begin{tabular}{|l|l|l|l|l|l|l|l|l|l|}
			\hline
			Backbones & $\times +$  & $P$ &Acc    & AP & AP$_{50}$ & AP$_{75}$ & AP$_S$ & AP$_M$ & AP$_L$ \\
			& (M) & (M) & (\%) &(\%) & (\%)& (\%)&(\%) &(\%) &(\%) 
			\\
			\hline\hline
			MobileNetV3 & 219 & 5.4 & 75.2& 29.9 & 49.3 & 30.8 & 14.9 & 33.3 & 41.1\\
			MnasNet-A2 & 340& 4.8 & 75.6 & 30.5 & 50.2 & 32.0 & 16.6 & 34.1 & 41.1\\
			SingPath NAS & 365 & 4.3 & 75.0 & 30.7 & 49.8 & 32.2 & 15.4 &33.9 & 41.6\\
			MixNet-M & 360 & 5.0 & 77.0 & 31.3& 51.7 & 32.4& 17.0 & 35.0 & 41.9   \\
			MixPath-A & 349& 5.0 & 76.9 &31.5& 51.3 & 33.2 & 17.4& 35.3& 41.8\\
			\hline
		\end{tabular}
	\end{center}
	\caption{COCO Object detection with various drop-in backbones}
	\label{table:mixpath-coco-retina}
\end{table}

\subsection{Comparison of search strategies}

\begin{figure}[ht]
	\centering
	\includegraphics[scale=0.6]{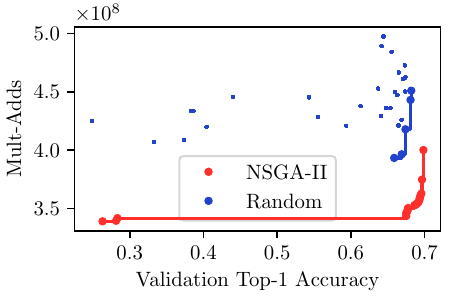}
	\caption{Pareto-front of models by NSGA-II vs. random search.}
	\label{fig:nsga-vs-random}
\end{figure}

We show the Pareto front of models searched by NSGA-II vs. Random in Fig.~\ref{fig:nsga-vs-random}.

\subsection{More statistics analysis on SBNs}\label{subsec:bn-m-geq3}

\begin{figure*}[tb!]
	\centering
	\begin{subfigure}{\linewidth}
		\includegraphics[scale=0.6]{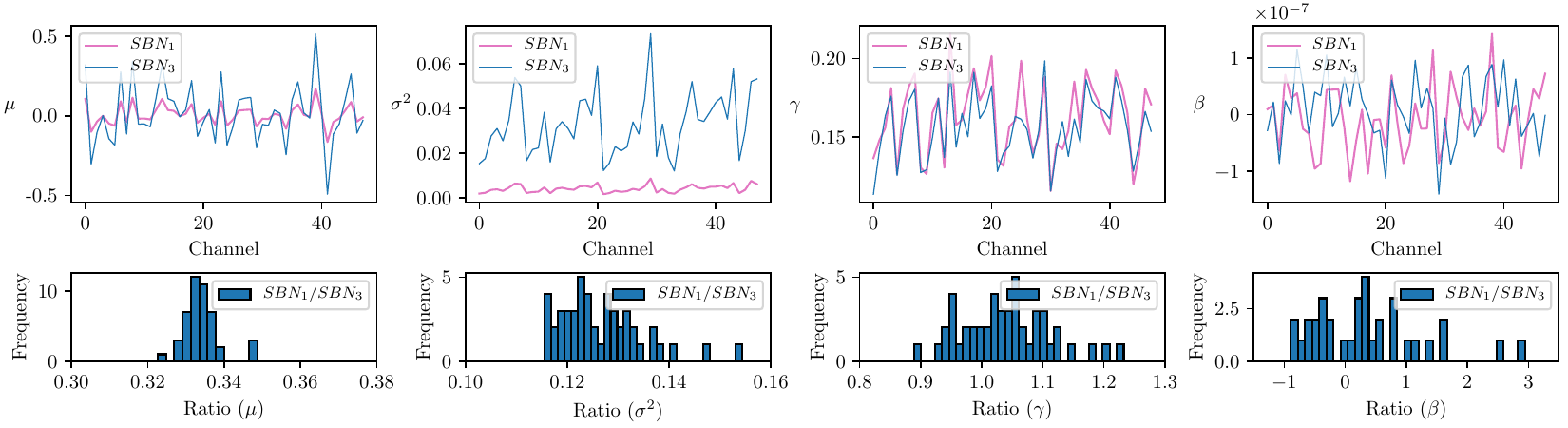}
		\caption{$m=3$}
	\end{subfigure}
	\begin{subfigure}{\linewidth}
		\includegraphics[scale=0.6]{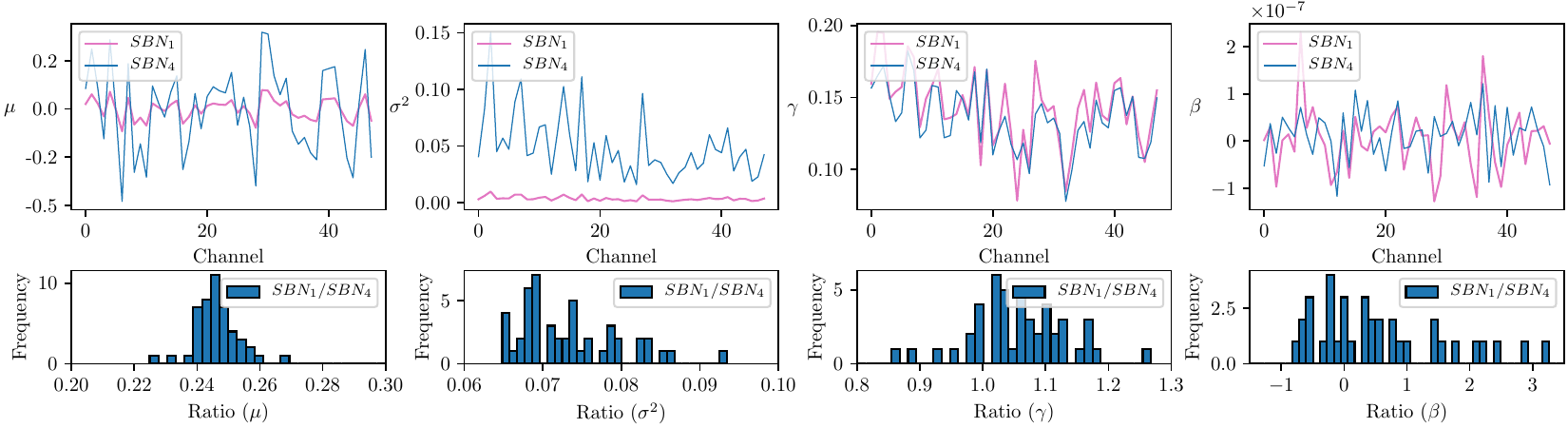}
		\caption{$m=4$}
	\end{subfigure}
	\caption{Parameters ($\mu, \sigma^2, \gamma, \beta$) of the first-layer SBNs in MixPath supernet (in $S_1$) trained on CIFAR-10  when at most $m=3,4$ paths can be activated. $SBN_n$ refers to the one follows $n$-path activations. The parameters of $SBN_3$ and $SBN_4$ are multiples of $SBN_1$ as expected.}
	\label{fig:sbn-relation-m3-m4}
\end{figure*}

\begin{figure*}[tb!]
	\centering
	\begin{subfigure}{\linewidth}
		\includegraphics[scale=0.6]{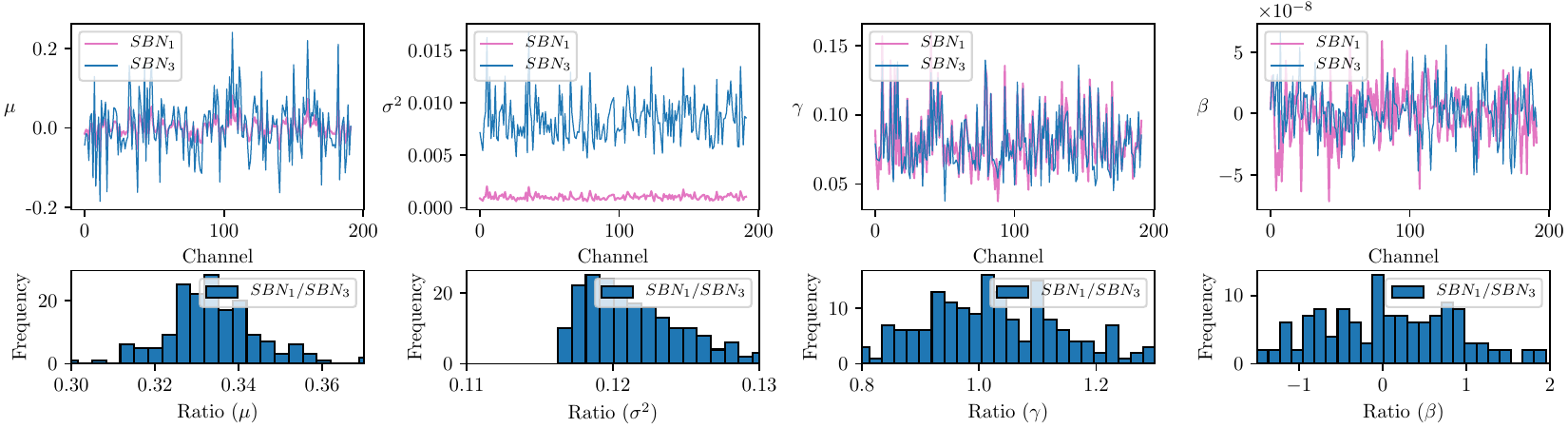} 
		\caption{Layer 6}
	\end{subfigure}
	\begin{subfigure}{\linewidth}
		\includegraphics[scale=0.6]{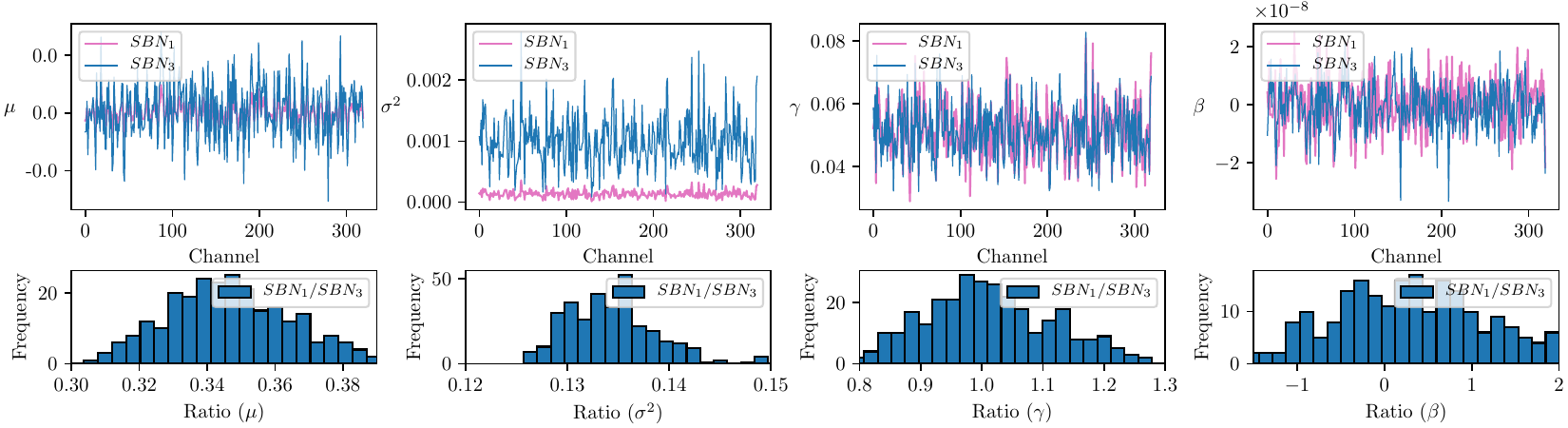} 
		\caption{Layer 11}
	\end{subfigure}
	\caption{Parameters ($\mu, \sigma^2, \gamma, \beta$) of the SBNs of the 6th and 11th layer in MixPath supernet (in $S_1$) trained on CIFAR-10 when at most $m=3$ paths can be activated. $SBN_n$ refers to the one follows $n$-path activations. The parameters of $SBN_3$ are still multiples of $SBN_1$ as expected.}
	\label{fig:sbn-relation-m3-l6-l11}
\end{figure*}

To further confirm our early postulation, we train MixPath supernet in the search space $S_1$ on CIFAR-10, allowing the number of activable paths $m=3$ and $m=4$. Other settings are kept the same as the case $m=2$. The relationship of parameters in SBNs is shown in Fig.~\ref{fig:sbn-relation-m3-m4}. As expected, SBNs capture feature statistics for different combinations of path activation. For instance, the mean of $SBN_3$ is three times that of $SBN_1$. The similar phenomenon can be observed in other layers as well, for instance, the statistics of the 6-th and 11-th layer are shown in Fig.~\ref{fig:sbn-relation-m3-l6-l11}. 

\begin{figure}
	\centering
	\includegraphics[scale=0.45]{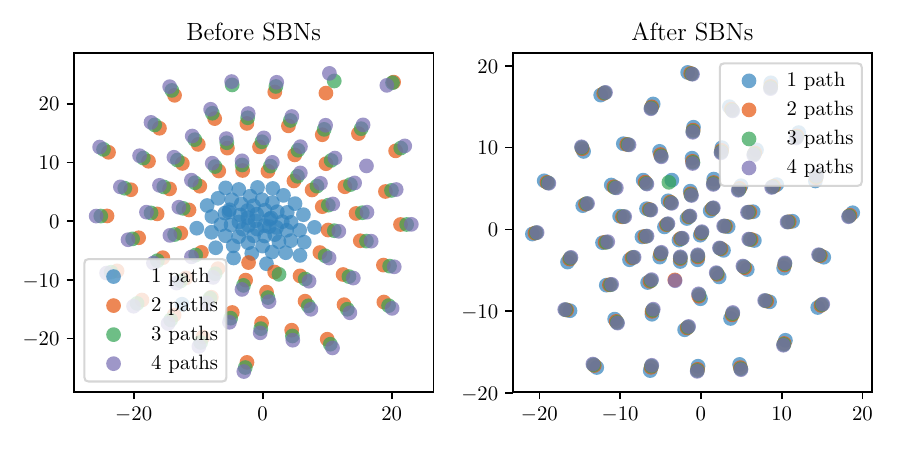}
	\caption{A t-SNE visualization of first-layer multi-path features before and after SBNs. We randomly sample 64 samples to get these features. Dots of the same color indicate the same multi-path combination. SBNs make distant features from multi-path combinations similar to each other (see closely overlapped dots on the right). Best viewed in color.}
	\label{fig:tsne_multiple_path}
\end{figure}

\textbf{SBNs have a strong impact on regularizing features from multiple paths\quad} This is more obvious when we draw a t-SNE  visualization \cite{van2008visualizing} of first-layer feature maps from our MixPath supernet ($m=4$) trained on CIFAR-10 in Fig.~\ref{fig:tsne_multiple_path}. Before applying SBNs, features from different path combinations are quite distant from each other, while SBNs close up this gap and make them quite similar.

\subsection{Cosine similarity and feature vectors on NAS-Bench-101}
\begin{figure*}[tb!]
	\centering
	\begin{subfigure}{0.6\linewidth}
		\includegraphics[width=\textwidth,scale=0.55]{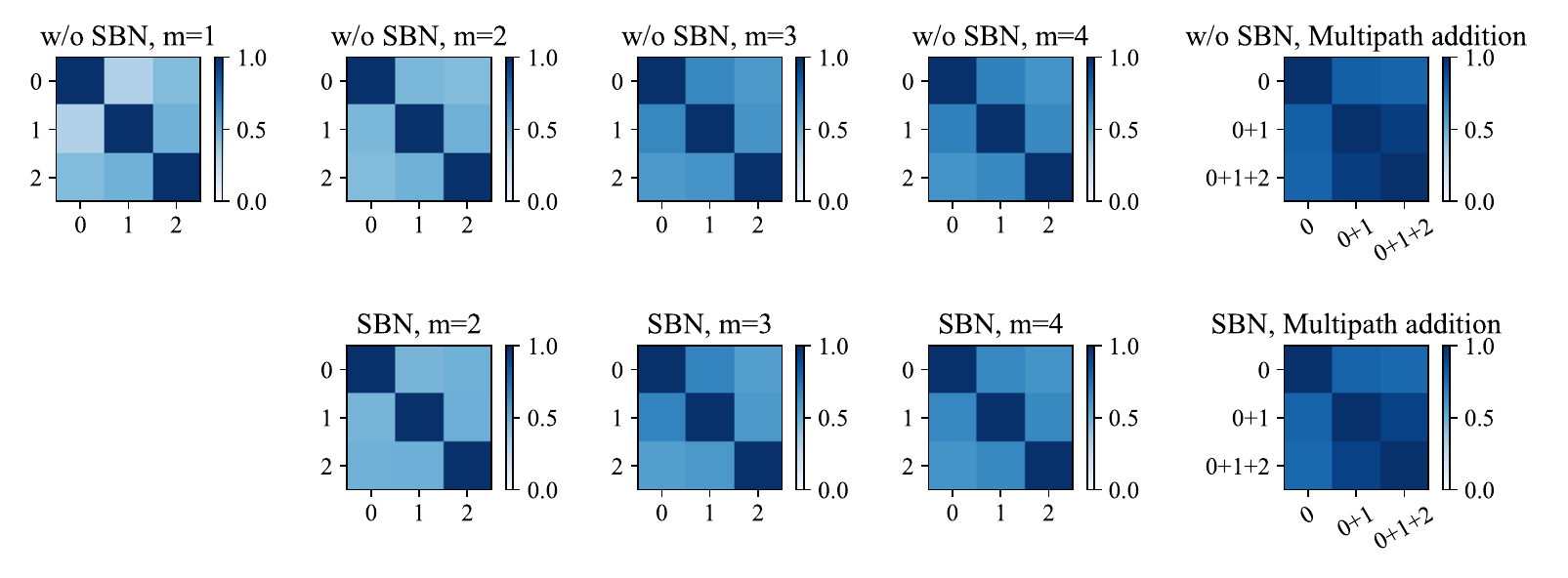}
	\end{subfigure}
	\begin{subfigure}{0.35\linewidth}
		\includegraphics[width=0.5\textwidth,scale=0.55]{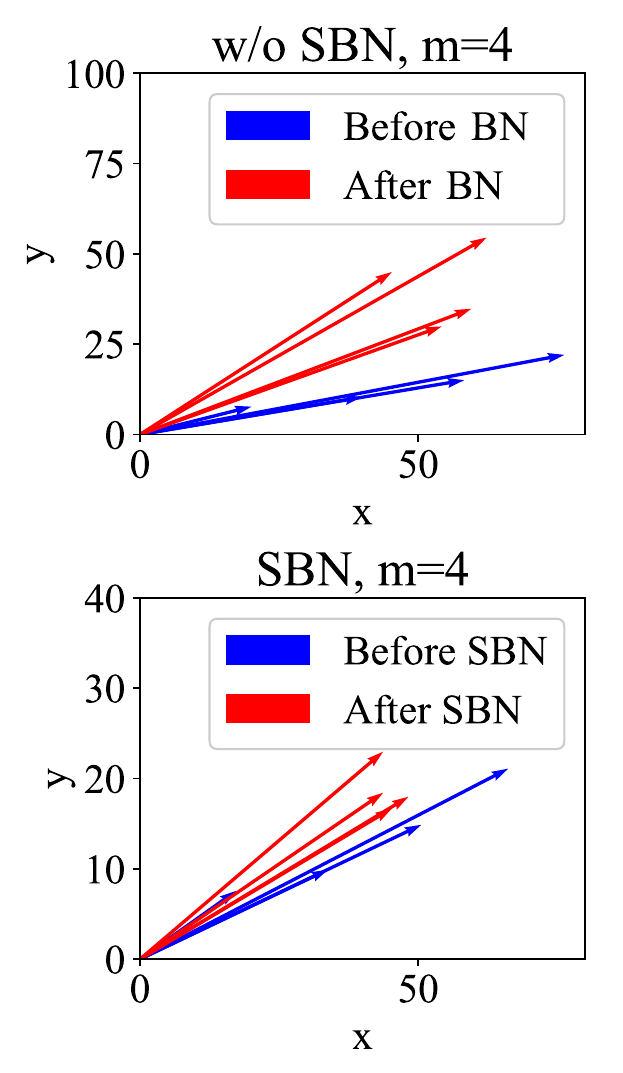}
	\end{subfigure}
	\caption{\textbf{(a)} Cosine similarity of first-block features from the supernets trained on NAS-Bench-101  with and without SBNs \textbf{(b)} Feature vectors projected into 2-dimensional space.}
	\label{fig:cosine1}
\end{figure*}

We also plot the cosine similarity of features from different operations along with their projected vectors before/after SBNs and vanilla BNs on NAS-Bench-101 in Fig.~\ref{fig:cosine1}. We can see that not only are the features from different operations similar, but so are the summations of features from multiple paths. At the same time, SBNs can transform the amplitudes of different vectors to the same level, while vanilla BNs can't. This is similar to the situation in the search space $S_{1}$ and matches with our theoretical analysis.

\begin{figure*}[tb!]
	\centering
	\includegraphics[scale=0.15]{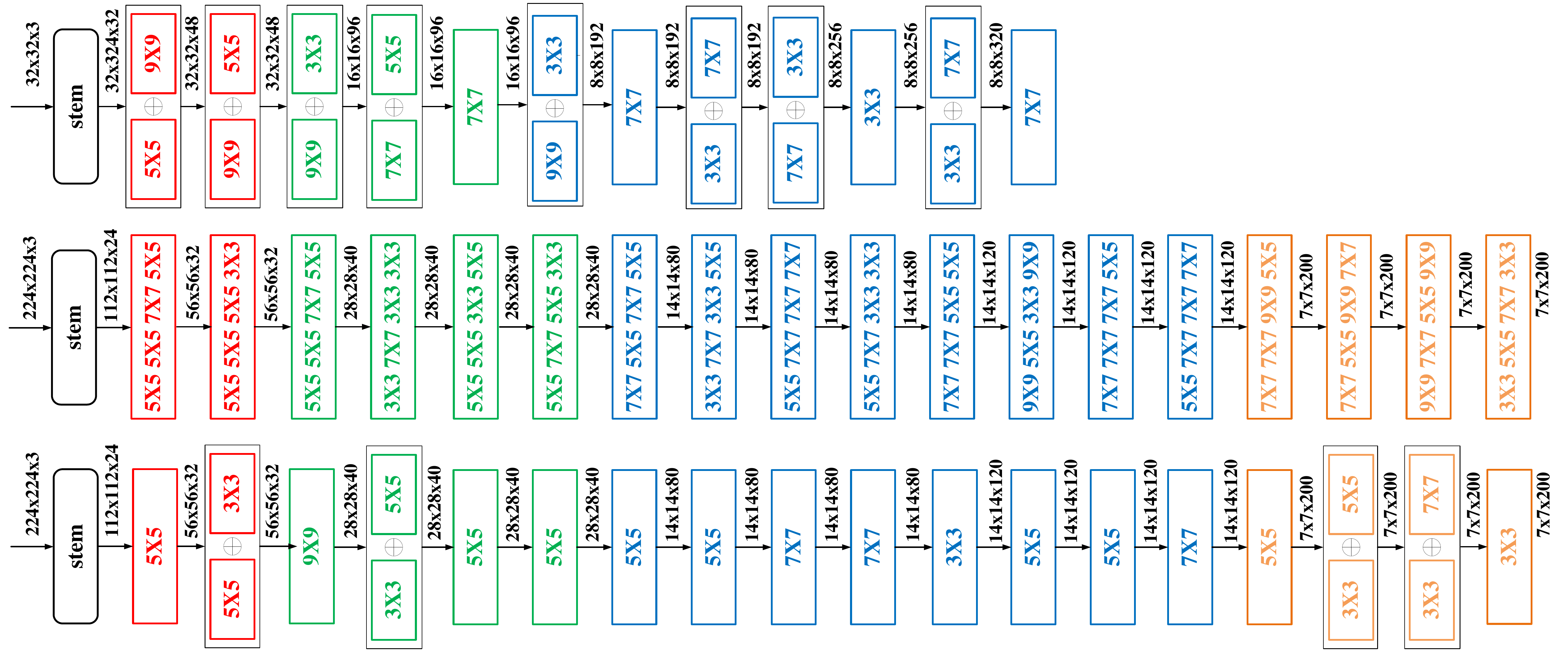}
	\caption{The architecture of MixPath-c (top), MixPath-A (middle) and MixPath-B (bottom). MixPath-B makes use of feature aggregation and outperforms EfficientNet-B0 with fewer FLOPS and parameters.}
	\label{fig:mpos-architecture}
\end{figure*}

\subsection{Searched architectures on CIFAR-10 and ImageNet}\label{subsec:architecture}
The architectures of MixPath-c, MixPath-A and MixPath-B are shown in Fig~\ref*{fig:mpos-architecture}.

\end{document}